\documentclass[11pt,letterpaper]{article}
\usepackage{emnlp2017}
\usepackage{times}
\usepackage{latexsym}

\emnlpfinalcopy

\usepackage{fancyvrb}
\usepackage{float}
\usepackage{url}
\usepackage{amsmath}  
\usepackage{amsfonts} 
\usepackage{tikz-qtree-compat}
\usepackage{forest}
\usepackage[e]{esvect} 
\usepackage{subcaption} 
\usepackage{relsize} 
\usetikzlibrary{arrows} 
\usepackage{enumitem}
\usepackage{todonotes} 
\usepackage{tcolorbox} 

\def\emnlppaperid{46}

\newcommand{\langpro}{\textsc{l}ang\textsc{p}ro}
\newcommand{\llfgen}{\textsc{llf}gen}
\newcommand{\nlogpro}{\textsc{nl}og\textsc{p}ro}

\newcommand{\X}{\mathbb{X}}
\newcommand{\T}{\mathbb{T}}
\newcommand{\F}{\mathbb{F}}

\newcommand{\BS}{\backslash}
\newcommand{\NP}{N\!P}
\newcommand{\VP}{{V\!P}}
\newcommand{\PP}{{P\!P}}

\newcommand{\pp}{{\tt pp}}
\newcommand{\np}{{\tt np}}
\newcommand{\sen}{{\tt s}}
\newcommand{\vp}{{\tt vp}}
\newcommand{\qu}{{\tt q}}
\newcommand{\nou}{{\tt n}}

\newcommand{\semt}[1]{\mathtt{#1}}
\newcommand{\synt}[1]{\textbf{\textsmaller[1]{#1}}}

\newcommand{\btimes}{{\scalebox{1.15}{$\times$}\kern-1.5pt}}
\newcommand{\ndList}[1]{\scalebox{.85}{[#1]}}
\newcommand{\abst}[1]{\lambda #1.\,}

\newcommand{\ruleConst}[1]{\textcolor{gray}{#1}}
\newcommand{\subs}{\sqsubseteq}

\newcommand{\toppad}[1]{\rule{0pt}{#1}}

\newcommand{\addlater}[1]{}

\newtcbox{\lab}[1][white]{on line, arc=1.5pt, colframe=black!30, colback=#1, 
	boxrule=1pt, boxsep=0pt, left=1.5pt, right=1.5pt, top=1.5pt, bottom=1.5pt}

\newcommand{\nonBranchingRule}[3][]{
\begin{tabular}{@{}c@{}}
$\dfrac{\begin{tabular}{@{}c@{}}#2\end{tabular}}
{\raisebox{-4pt}{\begin{tabular}[t]{@{}c@{}}#3\end{tabular}}}$#1
\end{tabular}}

\newcommand{\BranchingRule}[4][]{
\begin{tabular}{@{}c@{}}
$\dfrac{\begin{tabular}{@{}c@{}}#2\end{tabular}}
{\raisebox{-4pt}{\begin{tabular}[t]{@{}c}#3\end{tabular}} ~~
\raisebox{-4pt}{\begin{tabular}[t]{c@{}}#4\end{tabular}}}$#1
\end{tabular}}

\forestset{
  labA/.style n args=3{edge label/.expanded={node [midway, auto=#1, font=\unexpanded{#2}]{\unexpanded{#3}}},
  },
  labB/.style n args=3{label={[label distance=#1, font=\unexpanded{#2}]-90:{\unexpanded{#3}}},
  },
  labelA/.style ={edge label/.expanded={node [midway, auto=right, font=\unexpanded{\footnotesize}]{\unexpanded{#1}}},
  },
  labelB/.style ={label={[label distance=3pt, font=\unexpanded{\footnotesize}]-90:{\unexpanded{#1}}},
  }
}

\makeatletter
\newcommand{\verbatimfont}[1]{\def\verbatim@font{#1}}%
\makeatother

\newcommand{\comments}[1]{}

\newfloat{mycode}{tbhp}{lst}
\floatname{mycode}{Code}

\allowdisplaybreaks 

\title{\textsc{langpro}: Natural Language Theorem Prover}

\author{Lasha Abzianidze\\
  CLCG, University of Groningen\\The Netherlands\\
  {\tt L.Abzianidze@rug.nl}}

\date{}

\begin{document}
\maketitle

\begin{abstract}

\langpro{} is an automated theorem prover for natural language.%
\footnote{\url{https://github.com/kovvalsky/LangPro}
}
Given a set of premises and a hypothesis, it is able to prove semantic relations between them.
The prover is based on a version of analytic tableau method specially designed for natural logic.
The proof procedure operates on logical forms that preserve linguistic expressions to a large extent. 
The nature of proofs is deductive and transparent.
On the FraCaS and SICK textual entailment datasets, the prover achieves high results comparable to state-of-the-art.

\end{abstract}

\section{Introduction}

Nowadays many formal logics come with their own proof systems and with the automated theorem provers based on these systems.
If we share Montagues's famous belief
that there is ``no important theoretical difference between natural languages and the artificial languages of logicians'', then there plausibly exists a proof system for natural languages too.
On the other hand, studies on Natural Logic seek a formal logic whose formulas are as close as possible to linguistic expressions.    
Inspired by these research ideas, \citet{muskens:10} proposed an analytic tableau system for natural logic, where higher-order logic based on a simple type theory 
is used as natural logic and a version of analytic tableau method is designed for it.  
Later, \citet{abzianidze:2015:LENLS,abzianidze:2015:EMNLP,abzianidzethesis} made the tableau system suitable for wide-coverage reasoning by extending it and implementing a theorem prover based on it.



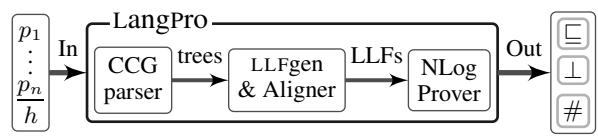
\begin{figure}[t]
\centering
\begin{tikzpicture}[scale=1]
\def\bgcol{gray!15}
\def\linecol{black!70}

\tikzset{every path/.style={line width=2pt}}
\tikzset{every node/.style={font=\small,scale=1, align=center, rounded corners=2pt, rectangle, draw=none, line width=.5pt, inner sep=1.2mm, outer sep=0pt}}

\node(langpro)at(0,0)[draw=black!90,line width=1pt, minimum height=13mm, minimum width=54mm, 
label={[anchor=west,fill=white,inner sep=1pt,shift={($(langpro.north west)+(.3,0)$)}]90:{\normalsize \langpro{}}}]{};
\node(parser)at($(langpro.west)+(.1,-.1)$)[draw=\linecol,anchor=west]{CCG\\parser};
\node(llfgen)at($(parser.east)+(.75,0)$)[anchor=west, draw=\linecol]{\llfgen{}\\\& Aligner};
\node(natpro)at($(langpro.east)+(-.1,-.1)$)[draw=\linecol,anchor=east]{NLog\\Prover};

\node(argument)at($(langpro.west)+(-.5,0)$)[draw=\linecol,anchor=east,inner ysep=0pt,inner xsep=2pt]{
\begin{tabular}{@{}c@{}}
$p_1$\toppad{3mm}\\[-1.4mm] \vdots\\[-1.2mm] $p_n$\\\hline $h$\toppad{3mm}\\
\end{tabular}};

\node(output)at($(langpro.east)+(.7,0)$)[draw=\linecol,anchor=west,inner ysep=2pt,inner xsep=2pt]{
\begin{tabular}{@{}c@{}}
\lab{$\subs$}\\[1mm] \lab{$\bot$}\\[1mm] \lab{$\#$}
\end{tabular}};

\path [arrows={-latex'}] (parser.east) edge[draw=\linecol,line width=1.5pt] node[sloped,above,outer sep=1mm,shift={(0mm,0mm)}]{\footnotesize trees} (llfgen.west);  

\path [->,>=latex'] (llfgen.east) edge[draw=\linecol,line width=1.5pt] node[sloped,above,outer sep=1mm]{\footnotesize LLFs} (natpro.west); 

\path [->,>=latex'] (argument.east) edge[draw=\linecol] node[sloped,above,outer sep=1mm]{\footnotesize In} (langpro.west);

\path [->,>=latex'] (langpro.east) edge[draw=\linecol] node[sloped,above,outer sep=1mm]{\footnotesize Out} (output.west);

\end{tikzpicture}
\caption[]{\langpro{} checks whether a set of premises $p_1,\ldots,p_n$ entails ($\subs$), contradicts ($\bot$) or is neutral ($\#$) to a hypothesis $h$.}
\label{fig:langPro_lean} 
\end{figure}

This paper presents the Prolog implementation of the theorem prover, called \langpro{},
%
in detail and completes the previous publications in terms of the system description.    
The rest of the paper is organized as follows.
First, we briefly introduce the tableau system and the employed natural logic. 
Then we characterize the architecture and functionality of \langpro{} (see Figure\,\ref{fig:langPro_lean}).
Before concluding, we briefly compare the prover to the related textual entailment systems.


\section{Natural Tableau}
\label{sec:TSNL}

\begin{figure}[t!]
\centerline{
\scalebox{.8}{
\begin{forest}
for tree={	align=center, 
			parent anchor=south, 
			child anchor=north, 
			l sep=4mm,
			s sep=4mm}			
[{\lab{1}~$\synt{several}~\synt{pug}~\synt{bark} : \T$}\\
 {\lab{2}~$\synt{every}(\synt{which}~\synt{bark}~\synt{dog})(\synt{be}~\synt{vicious}) : \T$}\\
 {\lab{3}~$\synt{no}~\synt{pug}(\synt{be}~\synt{evil}) : \T$}
 [{\lab{4}~$\synt{pug} \!:\! c \!:\! \T$}\\
  {\lab{5}~$\synt{bark} \!:\! c \!:\! \T$}
  ,labelA={$\exists_\T$\ndList{1}} 
  ,labB={1pt}{\footnotesize}{$\forall_\T$\ndList{2}} 
  [{\lab{6}~$\synt{which}~\synt{bark}~\synt{dog} \!:\! c \!:\! \F$} 
   ,labB={1pt}{\footnotesize}{$\wedge_\F$\ndList{6}} 
   [{\lab{8}~$\synt{bark} \!:\! c \!:\! \F$}
    [{\lab{10}~$\btimes$}
     ,labelA={$\btimes$\ndList{5,8}} 
    ]
   ]
   [{\lab{9}~$\synt{dog} \!:\! c \!:\! \F$}
     [{\lab{11}~$\btimes$}
      ,labelA={$\btimes$\ndList{4,9}} 
     ]
   ]
  ]
  [{\lab{7}~$\synt{be~vicious} \!:\! c \!:\! \T$}
   [{\lab{12}~$\synt{vicious} \!:\! c \!:\! \T$}
    ,labelA={$\textsc{aux}$\ndList{7}}
    ,labB={1pt}{\footnotesize}{$\textsc{no}_\T$\ndList{3}} 
    [{\lab{13}~$\synt{pug} \!:\! c \!:\! \F$}
     [{\lab{15}~$\btimes$}
      ,labelA={$\btimes$\ndList{4,13}} 
     ]
    ] 
    [{\lab{14}~$\synt{be~evil} \!:\! c \!:\! \F$}
     [{\lab{16}~$\synt{evil} \!:\! c \!:\! \F$}
      ,labelA={$\textsc{aux}$\ndList{14}}
      [{\lab{17}~$\btimes$}
      ,labelA={$\btimes$\ndList{12,16}} 
      ]
     ]
    ] 
   ] 
  ]
 ] 
]  
\end{forest}
}
}
\caption{The tableau proves: {\em several pugs bark}. {\em every dog which barks is vicious}. $\bot$ {\em no pug is evil}.}
\label{fig:pug}  
\end{figure}
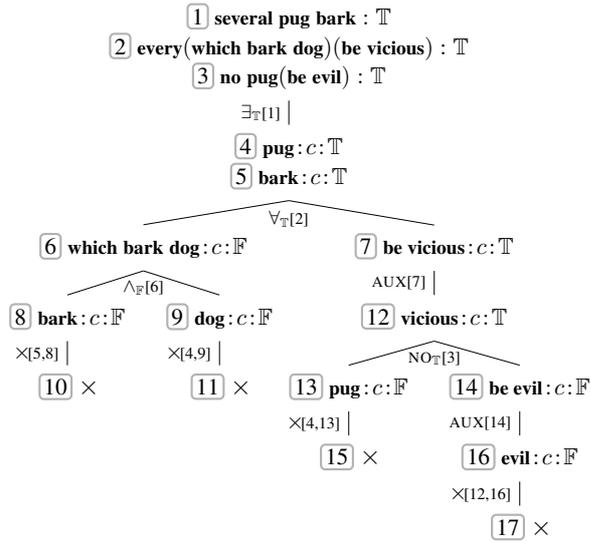

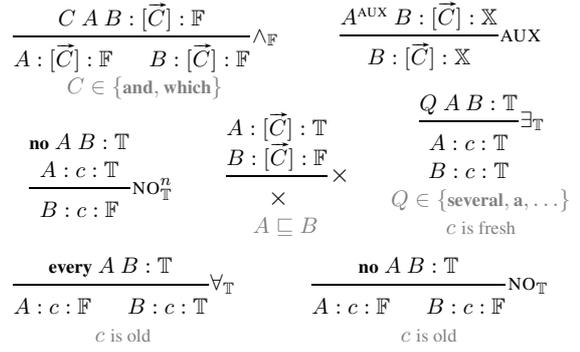
\begin{figure}[t]
\centering
\begin{tikzpicture}
\tikzset{every node/.style={scale=.77}}

\node(and_f) at(0,0) [anchor=west]  {
	\BranchingRule[$\wedge_\F$\\\ruleConst{$C\in \{\synt{and},\synt{which}\}$}]
	{$C~A~B : [\vv{C}] : \F$}
	{$A : [\vv{C}] : \F$}{$B : [\vv{C}] : \F$}
};
\node(aux) at($(and_f.east)+(.5,.2)$)[anchor=west] {
	\nonBranchingRule[\textsc{aux}]
	{$A^{\textsc{aux}} ~B : [\vv{C}] : \X$}
	{$B : [\vv{C}] : \X$}
};

\node(ex_t) at(5,-1.5) [anchor=west]  {
	\nonBranchingRule[$\exists_\T$\\\ruleConst{$Q \in \{\synt{several},\synt{a},\ldots\}$}\\\ruleConst{$c$ {\small is fresh}}]
	{$Q~A~B : \T$}
	{$A : c : \T$\\$B : c : \T$}
};
\node(cl) at($(ex_t.west)+(-.3,-.15)$)[anchor=east] {
	\nonBranchingRule[$\btimes$\\\ruleConst{$A\subs B$}]
	{$A : [\vv{C}] : \T$\\$B : [\vv{C}] : \F$}
	{$\btimes$}
};
\node(no_n_t) at($(cl.west)+(-.4,0)$)[anchor=east] {
	\nonBranchingRule[$\textsc{no}^n_\T$]
	{$\synt{no}~A~B : \T$\\$A : c : \T$}
	{$B : c : \F$}
};

\node(all_t) at(0,-3.3) [anchor=west]  {
	\BranchingRule[$\forall_\T$\\\ruleConst{$c$ {\small is old}}]
	{$\synt{every}~A~B : \T$}
	{$A : c : \F$}{$B : c : \T$}
};
\node(no_t) at($(all_t.east)+(.7,0)$) [anchor=west] {
	\BranchingRule[$\textsc{no}_\T$\\\ruleConst{$c$ {\small is old}}]
	{$\synt{no}~A~B : \T$}
	{$A : c : \F$}{$B : c : \F$}
};
\end{tikzpicture}
\caption{The inference rules employed in the tableau proof of Figure\,\ref{fig:pug}. 
An entity term is {\em old} ({\em fresh}) wrt a branch iff it is (not) in the branch.}
\label{fig:rules}
\end{figure}

An {\em analytic tableau method} 
is a proof procedure which searches a model, i.e. a possible situation, satisfying a set of logic formulas.
The search is performed by gradually applying inference rules, also called {\em tableau rules}, to the formulas. 
A~tableau rule has antecedents and consequent and is easy to read, e.g., according to $\textsc{no}_\T$ in Figure\,\ref{fig:rules}, if no $A$ is $B$, then for any entity $c$, either it is not $A$ or it is not $B$.
A~tableau proof, in short a {\em tableau}, is often depicted as an upside-down tree with initial formulas at its root (Figure\,\ref{fig:pug}).
After each rule application, new inferred formulas are introduced in the tableau.
Depending on the applied rules, the tableau can branch or grow in depth.
A tableau branch models a situation that satisfies all the formulas in the branch.
Closed branches, marked with $\btimes$, correspond to inconsistent situations.
The search for a possible situation fails if all branches are closed---the tableau is closed. 

The natural tableau is a tableau method for a version of natural logic.%
\footnote{It is an extended version of Muskens' original  tableau system.
The extension is three-fold and concerns the type system, the format of tableau entries and the inventory of tableau rules \cite{abzianidze:2015:LENLS}.
}
The terms of the natural logic, called {\em Lambda Logical Forms} (LLFs), are simply typed $\lambda$-terms built up from variables and constant lexical terms with the help of function application and $\lambda$-abstraction. 
The format of a tableau entry, i.e. node, is a tuple consisting of a modifier list, an LLF, an argument list and a truth sign. The parts are delimited with a colon.
The empty lists are omitted for conciseness.
For example, the entries \eqref{eq:loudly_bark:1} and \eqref{eq:loudly_bark:2} both mean that it is true that $c$ barks loudly in Paris, where $(\alpha,\beta)$ is a functional type that expects an argument of type $\alpha$ and returns a value of type $\beta$.%
\footnote{LLFs are typed with syntactic and semantic types.
Interaction between these types is established via the subtyping relation, e.g., entities being a subtype of NPs, $e<:\np$, makes $\synt{bark}_\vp~c_e$ well-formed, where
$\vp$ abbreviates $(\np,\sen)$.       
} 
\begingroup
\abovedisplayskip=2mm\abovedisplayshortskip=1mm
\belowdisplayskip=2mm\belowdisplayshortskip=1mm
\begin{align}
\synt{in}_{\np,\vp,\vp} \synt{Paris}_\np : \synt{loudly}_{\vp,\vp} \synt{bark}_\vp : c_e : \T
\label{eq:loudly_bark:1}
\\
(\synt{in}_{\np,\vp,\vp} \synt{Paris}_\np) (\synt{loudly}_{\vp,\vp} \synt{bark}_\vp~c_e) : \T
\label{eq:loudly_bark:2}
\end{align}
\endgroup

In order to prove a certain logical relation between premises and a hypothesis, the natural tableau searches a situation for the counterexample of the relation.  
The relation is proved if the situation is not found, otherwise it is refuted.
An example of a closed natural tableau is shown in Figure\,\ref{fig:pug}.
It proves the contradiction relation as it fails to find a situation for the counterexample---the premises and the hypothesis being true.
In order to facilitate reading tableau proofs, type information is omitted, the entries are enumerated and arcs are labeled with tableau rule applications.
For example, \lab{4} and \lab{5} are obtained by applying $\exists_\T$ to \lab{1}: if it is true that several pugs bark, then there is some entity $c$ which is a pug and which barks.


\section{LLF Generator}
\label{sec:llfs}

\begin{figure}[t]
\centering
\begin{tikzpicture}
\def\bgcol{gray!15}
\def\linecol{black!50}

\tikzset{every path/.style={line width=1.5pt}}
\tikzset{every node/.style={font=\small,line width=1pt, inner sep=3pt, outer sep=0pt, align=center, draw, rounded corners=1mm}}

\node(llfgen)at(0,0)[anchor=west, minimum height=12mm, minimum width=54mm, label={[anchor=west,fill=white,inner sep=1pt,shift={($(llfgen.north west)+(.3,-.05)$)}]90:{\normalsize \llfgen{}}}]{}; 

\node(ccgtree)at($(llfgen.west) + (-.4,0)$)[fill=white, anchor=west, draw=\linecol]
{CCG\\Tree};   

\node(ccgterm)at($(ccgtree.east) + (.7,0)$)[fill=white,anchor=west, draw=\linecol]
{CCG\\Term}; 

\node(ccgfixedterm)at($(ccgterm.east) + (.7,0)$)[anchor=west, draw=\linecol]
{Corrected\\CCG Term}; 

\node(llf)at($(ccgfixedterm.east) + (.7,0)$)[fill=white,anchor=west, draw=\linecol]
{LLFs}; 

\node(sr1)at($(llf.east) + (.5,.4)$)[black!40, anchor=west]
{FOL}; 

\node(sr2)at($(llf.east) + (.5,-.6)$)[black!40, anchor=west]
{DRT};  

\node(directionality)at($(ccgtree.east)+(0,-.9)$)[ line width=.3pt,font=\footnotesize, inner sep=2pt, fill=\bgcol, rectangle callout, callout pointer width=2mm, callout absolute pointer={($(ccgtree.east)+(.3,0)$)}]{Removing\\directionality}; 

\node(correcting)at($(ccgterm.east)+(.4,-.9)$)[line width=.3pt, font=\footnotesize, inner sep=3pt, fill=\bgcol, rectangle callout, callout pointer width=2mm, callout absolute pointer={($(ccgterm.east)+(.3,0)$)}]{Correcting\\analyses}; 

\node(typeraise)at($(ccgfixedterm.east)+(.2,-.9)$)[line width=.3pt, font=\footnotesize, inner sep=3pt, fill=\bgcol, rectangle callout, callout pointer width=2mm, callout absolute pointer={($(ccgfixedterm.east)+(.3,0)$)}]{Type-raising\\quantified NPs}; 

\path [->,>=latex'] (ccgtree.east) edge[\linecol] (ccgterm.west);

\path [->,>=latex'] (ccgterm.east) edge[\linecol] (ccgfixedterm.west);

\path [->,>=latex'] (ccgfixedterm.east) edge[\linecol] (llf.west);

\path [black!30,dotted,->,>=latex'] (llf.east) edge (sr1.west);

\path [black!30,dotted,->,>=latex'] (llf.east) edge (sr2.west);
\end{tikzpicture} 
\caption{The LLF generator produces a list of LLFs from a single CCG derivation tree.}
\label{fig:llf_gen}
\end{figure}
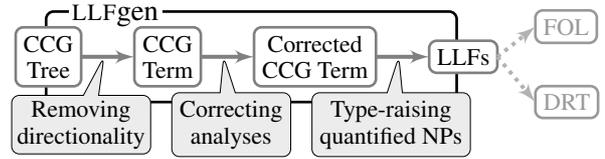

\begin{figure}[t]
\hspace{-3mm}\scalebox{.65}{
\begin{tikzpicture}[grow=down]
\tikzset{level distance = 25pt, sibling distance = 5pt}
\tikzset{every tree node/.style={align=center,anchor=north}}
\tikzset{level 2/.style={sibling distance=-40}}
\tikzset{level 1/.style={sibling distance=0}}
\Tree
[.\node{${\tt ba}[S_{dcl}]$};
 [.\node{\texttt{nobody}\\
   $\NP$\\
   \textbf{nobody}\\
   \normalsize{DT}}; 
 ]
 [.\node{${\tt fa}[\VP_{dcl}]$};
  [.\node{\texttt{is}\\
  	$\VP_{dcl} / \VP_{ng}$ \\
    \textbf{be}\\
    \normalsize{VBZ}}; 
  ]
  [.\node{${\tt fa}[\VP_{ng}]$};
   [.\node{${\tt fa}[\VP_{ng} / \PP]$};
    [.\node {\texttt{rinsing}\\
       $(\VP_{ng} / \PP)/\NP$ \\
       \textbf{rinse}\\
       \normalsize{VBG}}; 
    ]
    [.\node {${\tt fa}[\NP]$};
     [.\node{\texttt{a}\\
        $\NP/N$ \\
        \textbf{a}\\
        \normalsize{DT}}; 
     ]
     [.\node{\texttt{steak}\\
        $N$ \\
        \textbf{steak}\\
        \normalsize{NN}}; 
     ]
    ]
   ]
   [.\node{${\tt fa}[\PP]$};
    [.\node{\texttt{with}\\
      $\PP / \NP$ \\
      \textbf{with}\\
      \normalsize{IN}}; 
    ]
    [.\node{${\tt lx}[\NP, N]$};
     [.\node{\texttt{water}\\
       $N$ \\
       \textbf{water}\\
       \normalsize{NN}}; 
     ]
    ]
   ]
  ]
 ]
]
\end{tikzpicture}
}
\caption{The CCG tree by C\&C for {\em nobody~is rinsing a steak with water} (SICK-1379).
}
\label{ccg:1379}
\end{figure}
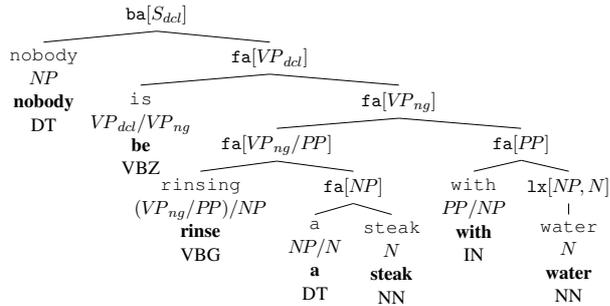

A Natural Tableau-based theorem prover for natural language requires automatic generation of LLFs from raw text.
To do so, we implement a module, called \llfgen{}, that generates LLFs from syntactic derivations of Combinatory Categorial Grammar (CCG, \citealt{Steedman:2000}).
Given a CCG derivation, \llfgen{} returns several LLFs that model different orders of quantifier scopes (see Figure\,\ref{fig:llf_gen}).%
\footnote{See \citet[Ch.\,3]{abzianidzethesis} for a detailed description.
}
Figure\,\ref{ccg:1379} displays a CCG derivation where $\VP_i$ abbreviates $S_i \BS \NP$.
%

\comments{
We generate LLFs from CCG derivations because (a) function-argument relations are readily available in Categorial Grammar-style derivations, and (b) CCG is the only Categorial Grammar which has robust syntactic parsers.
}  


LLFs are obtained from a CCG tree in three major steps (Figure\,\ref{fig:llf_gen}): (i) removing directionality from CCG trees, (ii) correcting semantically inadequate analyses, and (iii) type-raising quantified NPs (QNPs).
Below we briefly describe each of these steps and give corresponding examples.

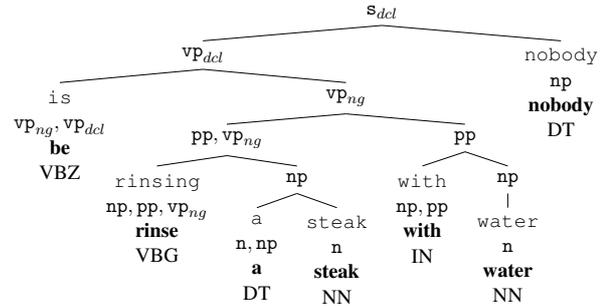
\begin{figure}[t]
\hspace{-3mm}
\scalebox{.7}{
\begin{tikzpicture}[grow=down]
\tikzset{level distance = 22.5pt, sibling distance = 8pt}
\tikzset{every tree node/.style={align=center,anchor=north}}
\tikzset{level 2/.style={sibling distance=-7}}
\tikzset{level 1/.style={sibling distance=-15}}
    \Tree
      [.\node{ $\sen_{dcl}$};
        [.\node{ $\vp_{dcl}$};
          [.\node{\texttt{is}\\
                  $\vp_{ng},\vp_{dcl}$ \\
                  \textbf{be}\\
                  \normalsize{VBZ}}; ]
          [.\node{ $\vp_{ng}$};
            [.\node{ $\pp,\vp_{ng}$};
              [.\node{\texttt{rinsing}\\
                      $\np,\pp,\vp_{ng}$ \\
                      \textbf{rinse}\\
                      \normalsize{VBG}}; ]
              [.\node{ $\np$};
                [.\node{\texttt{a}\\
                        $\nou,\np$ \\
                        \textbf{a}\\
                        \normalsize{DT}}; ]
                [.\node{\texttt{steak}\\
                        $\nou$ \\
                        \textbf{steak}\\
                        \normalsize{NN}}; ]
              ]
            ]
            [.\node{ $\pp$};
              [.\node{\texttt{with}\\
                      $\np,\pp$ \\
                      \textbf{with}\\
                      \normalsize{IN}}; ]
              [.\node{ $\np$};
                [.\node{\texttt{water}\\
                        $\nou$ \\
                        \textbf{water}\\
                        \normalsize{NN}}; ]
              ]
            ]
          ]
        ]
        [.\node{\texttt{nobody}\\
                $\np$ \\
                \textbf{nobody}\\
                \normalsize{DT}}; ]
      ]
  \end{tikzpicture}
}
\caption{The CCG term obtained from the CCG tree of Figure\,\ref{ccg:1379}.
NB: the lexical rule remains.}
\label{ccgterm:1379}
\end{figure}

Directionality information encoded in CCG categories and combinatory rules 
is redundant from a semantic perspective, therefore we discard it in the first step:  
CCG categories are converted into types ($Y\BS X$ and $Y/X$ $\rightsquigarrow$ $(\mathtt{x,y})$),
and argument constituents are placed after function ones in binary combinatory rules.
Resulted structures are called {\em CCG terms} (Figure\,\ref{ccgterm:1379}). 
%
\comments{
\begingroup
\abovedisplayskip=2mm\abovedisplayshortskip=1mm
\belowdisplayskip=2mm\belowdisplayshortskip=1mm
\begin{align}
((S_{ng}\BS\NP)/\PP)/\NP &\leadsto \np,\pp,\np,\sen_{ng}
\label{eq:rm_dir_1}
\\
{\tt bxc}[Z\!/\!X](\textbf{A}_{Y\!/\!X}, \textbf{F}_{Z\!\BS\!Y})
&\leadsto \abst{v} \textbf{F}_\semt{y,z} (\textbf{A}_\semt{x,y}~v_\semt{x}) 
\label{eq:rm_dir_2}
\end{align}
\endgroup
}

Obtained CCG terms are often semantically inadequate.\addlater{96\% of rules are eliminated}
One of the reasons for this is {\em lexical} (i.e. {\em type-changing}) rules (e.g., $N\!\mapsto\!\NP$ in Figure\,\ref{ccg:1379}) of the CCG parsers which still remain in CCG terms (e.g., $[\synt{water}_\nou]_\np$ in Figure\,\ref{ccgterm:1379}).
These rules are destructive from a compositional point of view.
We designed 13 schematic rewriting rules of general type that {\it correct} CCG terms---make them semantically more adequate and transparent.
The rules make use of types, part-of-speech (POS) and named entity (NE) tags to match semantically inadequate analyses:%
\footnote{
To handcraft the rules, we used a development set of 1.7K CCG derivations obtained by parsing the sentences from FraCaS \cite{fracas} and the trial portion of SICK \cite{sick:14} with CCG-based parsers: C\&C \cite{cc:2007} and EasyCCG \cite{lewis-steedman:14}.
}

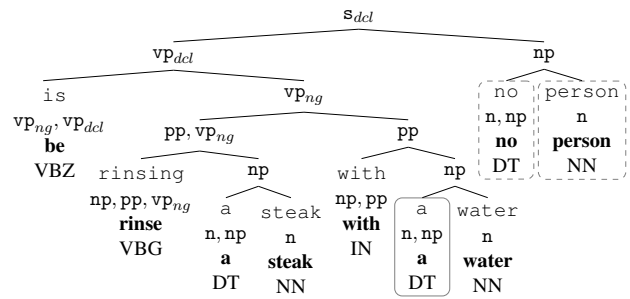
\begin{figure}[t]
\hspace{-4mm}
\scalebox{.70}{
\begin{tikzpicture}[grow=down]
\tikzset{level distance = 21pt, sibling distance=2pt}
\tikzset{every tree node/.style={align=center,anchor=north}}
\tikzset{level 5/.style={sibling distance=0}}
\tikzset{level 4/.style={sibling distance=0}}
\tikzset{level 3/.style={sibling distance=0}}
\tikzset{level 1/.style={sibling distance=-25}}
    \Tree
      [.\node{ $\sen_{dcl}$};
        [.\node{ $\vp_{dcl}$};
          [.\node {\texttt{is} \kern-15pt\\
                  $\vp_{ng},\vp_{dcl}$ \kern-20pt \\
                  \textbf{be} \kern-15pt\\
                  \normalsize{VBZ} \kern-15pt}; ]
          [.\node{ $\vp_{ng}$};
            [.\node{ $\pp,\vp_{ng}$};
              [.\node{\texttt{rinsing}\\
                      $\np,\pp,\vp_{ng}$ \\
                      \textbf{rinse}\\
                      \normalsize{VBG}}; ]
              [.\node{ $\np$};
                [.\node{\texttt{a}\\
                        $\nou,\np$ \\
                        \textbf{a}\\
                        \normalsize{DT}}; ]
                [.\node{\texttt{steak}\\
                        $\nou$ \\
                        \textbf{steak}\\
                        \normalsize{NN}}; ]
              ]
            ]
            [.\node{ $\pp$};
              [.\node{\texttt{with}\\
                      $\np,\pp$ \\
                      \textbf{with}\\
                      \normalsize{IN}}; ]
              [.\node{ $\np$};
                [.\node[rounded corners, draw=gray]{
                        \texttt{a}\\
                        $\nou,\np$ \\
                        \textbf{a}\\
                        \normalsize{DT}}; ]
                [.\node{\texttt{water}\\
                        $\nou$ \\
                        \textbf{water}\\
                        \normalsize{NN}}; ]
              ]
            ]
          ]
        ]
        [.\node{ $\np$};
          [.\node [dashed, rounded corners, draw=gray]{
          		  \texttt{no}\\
                  $\nou,\np$ \\
                  \textbf{no}\\
                  \normalsize{DT}}; ]
          [.\node [dashed, rounded corners, draw=gray]{
                  \texttt{person}\\
                  $\nou$ \\
                  \textbf{person}\\
                  \normalsize{NN}}; ]
        ]
      ]
  \end{tikzpicture}
}
\caption{The corrected version of the CCG term of Figure\,\ref{ccgterm:1379}, with inserted and decomposed terms.}
\label{fixccgterm:1379}
\end{figure}

\begin{itemize}[leftmargin=*]\itemsep0mm
\item 
Certain non-compositional multiword expressions are treated as constant terms:
{\em a lot of}, {\em in front of}, {\em a few}, {\em because of}, {\em next to}, etc.  

\item
Type-changing rules are {\em explained} by changing lexical types, decomposing terms or inserting new terms.
This step carries out conversions like 
$[\synt{europe}_\nou]_\np \rightsquigarrow \synt{europe}_\np$,
$[\synt{nobody}_\nou]_\np \rightsquigarrow \synt{no}_{\nou,\np} \synt{person}_\nou$, 
and $[\synt{water}_\nou]_\np \rightsquigarrow \synt{a}_{\nou,\np} \synt{water}_\nou$ (see Figure\,\ref{fixccgterm:1379}). 
Inserted $\synt{a}_{\nou,\np}$ 
merely plays a role of an existential 
quantifier.%
\addlater{Mention reduced relative clauses} 





\item 
Several CCG analyses are altered in order to reflect formal semantics, e.g., attributive modifiers are pushed {\it under} a relative clause:
$\synt{big}\,(\synt{which}~\synt{run}~\synt{mouse})$ 
$\rightsquigarrow$ $\synt{which}~\synt{run}\,(\synt{big}~\synt{mouse})$; and PPs are attached to nouns rather than NPs: 
$\synt{in}\,(\synt{a}~\synt{box}) (\synt{every}~\synt{pug})$ $\rightsquigarrow$
$\synt{every}\,(\synt{in}\,(\synt{a}~\synt{box})\,\synt{pug})$.

\comments{$\synt{big}_{\nou,\nou} (\synt{which}_{\vp,\nou,\nou} \synt{run}_\vp \synt{mouse}_n)$ 
$\rightsquigarrow\synt{which}_{\vp,\nou,\nou} \synt{run}_\vp (\synt{big}_{\nou,\nou} \synt{mouse}_n)$; and PPs are attached to nouns rather than NPs: 
$\synt{in}_{\np,\np,\np}(\synt{a}_{\nou,\np} \synt{box}_{\nou})(\synt{every}_{\nou,\np}\synt{pug}_{\nou})$
$\synt{every}_{\nou,\np}(\synt{in}_{\nou,\nou,\np}(\synt{a}_{\nou,\np} \synt{box}_{\nou})\synt{pug}_{\nou})$.
}


\comments{
\item Some phrases and lexical items are normalized.
For example, ${\synt{two}_{\pp,\nou} (\synt{of}_{\np,\pp} (\synt{the}_{\nou,\np} \synt{man}_\nou))}$ is normalized as 
$\synt{two}_{\nou,\np} \synt{man}_\nou$ 
while $\synt{each}_{\nou,\np}$ and $\synt{an}_{\nou,\np}$ are replaced by $\synt{every}_{\nou,\np}$ and $\synt{a}_{\nou,\np}$.
}

\end{itemize}

\comments{\begin{figure*}[b]
\hspace*{-7mm}\scalebox{.9}{
\parbox{18.5cm}{
\begin{align}
(\synt{no}_{n,\vp,\sen}~\synt{person}_\nou)
\big(\synt{be}_{\vp,\vp}
	\big(\lambda z~(\synt{a}_{n,\vp,\sen}~\synt{water}_\nou)
		\big(\lambda y~(\synt{a}_{n,\vp,\sen}~\synt{steak}_\nou)
			(\lambda x~\synt{rinse}_{\np,\pp,\vp}~x~
			(\synt{with}_{\np,\pp}~y)
			z)\big)\big)\big)
\\	
(\synt{no}_{n,\vp,\sen}~\synt{person}_\nou)
\big(\synt{be}_{\vp,\vp}
	\big(\lambda z~(\synt{a}_{n,\vp,\sen}~\synt{steak}_\nou)
		\big(\lambda x~(\synt{a}_{n,\vp,\sen}~\synt{water}_\nou)
			(\lambda y~\synt{rinse}_{\np,\pp,\vp}~x~
			(\synt{with}_{\np,\pp}~y)
			z)\big)\big)\big)\notag
\\			
(\synt{a}_{n,\vp,\sen}~\synt{water}_\nou)
\big(\lambda y~(\synt{a}_{n,\vp,\sen}~\synt{steak}_\nou)
	\big(\lambda x~(\synt{no}_{n,\vp,\sen}~\synt{person}_\nou)
		\big(\synt{be}_{\vp,\vp}~
		(\synt{rinse}_{\np,\pp,\vp}~x~
			(\synt{with}_{\np,\pp}~y)
			)\big)\big)\big)\notag
\\			
(\synt{a}_{n,\vp,\sen}~\synt{steak}_\nou)
\big(\lambda x~(\synt{a}_{n,\vp,\sen}~\synt{water}_\nou)
	\big(\lambda y~(\synt{no}_{n,\vp,\sen}~\synt{person}_\nou)
		\big(\synt{be}_{\vp,\vp}~
		(\synt{rinse}_{\np,\pp,\vp}~x~
			(\synt{with}_{\np,\pp}~y)
			)\big)\big)\big)	\notag						
\end{align}
}}
\end{figure*}}


LLFs are obtained from corrected CCG terms by type-raising QNPs from $\np$ to the type $(\vp,\sen)$ of generalized quantifiers.
Hence, several LLFs are produced from a single CCG tree due to quantifier scope ambiguity, e.g., (\ref{llf:nsw}--\ref{llf:wsn}) are some of the LLFs obtained from the CCG term of Figure\,\ref{fixccgterm:1379}.%
\footnote{The LLFs use the following abbreviations:
$S = \synt{a}_\qu~\synt{steak}_\nou$, 
$W = \synt{a}_\qu~\synt{water}_\nou$, 
and $N = \synt{no}_\qu~\synt{person}_\nou$, where $\qu = (\nou,\vp,\sen)$.
}

\parbox{72mm}{
\begingroup
\abovedisplayskip=0mm\abovedisplayshortskip=0mm
\belowdisplayskip=0mm\belowdisplayshortskip=0mm
\begin{align}
\kern-6mm
N \big(\synt{be}
	\big(\abst{z} S
		\big(\abst{x} W
			\big(\abst{y} \synt{rinse}\,x\,
			(\synt{with}\,y)
			z\big)\big)\big)\big)
\label{llf:nsw}			
\\\kern-6mm
N \big(\synt{be}
	\big(\abst{z} W
		\big(\abst{y} S
			\big(\abst{x} \synt{rinse}\,x\,
			(\synt{with}\,y)
			z\big)\big)\big)\big)
\label{llf:nws}			
\\\kern-6mm
W \big(\abst{y} S
	\big(\abst{x} N(\synt{be}\,\big(\synt{rinse}~x~
	(\synt{with}~y))\big)\big)\big)
\label{llf:wsn}
\end{align}
\endgroup   
}\medskip 

\comments{
\vspace{-7mm}
\begin{align*}
S &= \synt{a}_{\nou,\vp,\sen}~\synt{steak}_\nou 
& \synt{b} &= \synt{be}_{\vp,\vp} 
\\
W &= \synt{a}_{\nou,\vp,\sen}~\synt{water}_\nou 
& \synt{w} &= \synt{with}_{\np,\pp} 
\\
N &= \synt{no}_{\nou,\vp,\sen}~\synt{person}_\nou 
& \synt{r} &= \synt{rinse}_{\np,\pp,\vp}
\end{align*}
}

Since LLFs encode instructions for semantic composition, they can be used to compositionally derive  semantics in other meaning representations (Figure\,\ref{fig:llf_gen}), e.g., first-order logic (FOL) or Discourse Representation Theory (DRT).
For this application, \llfgen{} can be used as an independent tool.
Given a CCG derivation in the Prolog format (supported by both C\&C and EasyCCG), \llfgen{} can return LLFs in XML, HTML or \LaTeX formats.
For a CCG tree, it is also possible to get either only the first LLF, e.g., \eqref{llf:nsw}, often reflecting the natural order of quantifiers, or a list of LLFs with various quantifier scope orders (possibly including semantically equivalent LLFs, like \eqref{llf:nsw} and \eqref{llf:nws}).



\section{Natural Logic Theorem Prover}
\label{sec:NLogPro}

The tableau theorem prover for natural logic (\nlogpro{}) represents a core part of \langpro{} (Figure\,\ref{fig:langPro_lean}).
It is responsible for checking a set of linguistic expressions on (in)consistency.
\nlogpro{} consists of four components:
the {\em Proof Engine} builds tableau proofs by applying the rules from the {\em inventory of Rules};
the rule applications are validated by the properties of lexical terms (encoded in the {\em Signature}) and the lexical knowledge (available from the {\em Knowledge Base}).
We used the same development data for \llfgen{} and \nlogpro{}.

\subsection{Signature}
\label{sub:sig}

The signature (SG) lists lexical terms that have algebraic properties relevant for inference, e.g., monotonicity, intersectivity, and implicativity. 
The lexical items in the SG come with an argument structure where each argument position is associated with a set of algebraic properties.
For example, 
$\synt{every}$ is characterized in the SG as
%
\texttt{[dw,up]},
meaning that in its first argument $\synt{every}$ is downward monotone while being upward monotone in the second one.
Currently, the SG lists about 20 lexical items, mostly generalized quantifiers (GQs), that were found in the development data.\addlater{Check the number correctly}
%

\subsection{The Inventory of Rules}
\label{sub:rules}

The inventory of rules (IR) contains all inference rules used by the prover.
Currently there are ca. 80 rules in the IR (some in Figure\,\ref{fig:rules}).
Around a quarter of the rules are from \newcite{muskens:10} and the rest are manually collected while exploring the development data.
The rules cover a plethora of phenomena.
Some of them are of a formal nature like Boolean connectives and monotonicity and others of linguistic nature:  adjectives, prepositions, definite NPs, expletives, open compound nouns, light verbs, copula, passives and attitude verbs. 

The IR involves around 25 {\em derivable} rules---the rules that represent shortcuts of several rule applications.
One such rule is ($\textsc{no}^n_\T$) in Figure\,\ref{fig:rules}, which is a specific version of ($\textsc{no}_\T$). 
Use of derivable rules yields shorter tableau proofs but raises a problem of performing the same rule application several times.
\nlogpro{} avoids this by maintaining a subsumption relation between the rules and keeping track of rule applications per branch.

\addlater{how it looks like in reality?}


A user can introduce new rules in the IR as Prolog rules (Code\,\ref{code:rule}): the head of the rule encodes antecedent nodes \texttt{===>} consequent nodes, and the body is a list of Prolog goals specifying the conditions the rule has to meet.     

\begin{mycode}
\begin{Verbatim}[frame=single,fontsize=\footnotesize,tabsize=2,framerule=1pt]
r(Name, Feats, ConstIndx, KeyWrd, KB,
 br([nd(Mod1, LLF1, Arg1, Sign1),...
     nd(ModN, LLFN, ArgN, SignN)],
   Signature) ===>
 [br([nd(Mod3, LLF3, Arg3, Sign3),...], 
    Signature3),
  br([nd(Mod4, LLF4, Arg4, Sign4),...], 
    Signature4)]
:- Goal1, ..., GoalN. %conditions
\end{Verbatim}
\vspace{-6mm}
\caption{The Prolog format of tableau rules.
\texttt{Feats} denotes efficiency features, \texttt{ConstIndx} and \texttt{KB} are the KB and indexing of constants respectively (fixed for every rule), and \texttt{KeyWrd} denotes fixed lexical terms occurring in the rule.
Each branch maintains its own signature of entities introduced during the proof. 
}
\label{code:rule}
\end{mycode}

\subsection{Knowledge Base}
\label{sub:base}

The knowledge base (BS) is based on the Prolog version of WordNet 3.0 \cite{wordnet:98}.
At this moment only the hyponymy/hypernymy, similarity and antonymy relations are included in the KB.
For simplicity, \langpro{} does not do any word sense disambiguation (WSD)
%
but allows multiple word senses for a lexical term.
For example, $A \subs B$ iff $SynSet_A$ is a hyponym of $SynSet_B$, or there are similar $Sense_A$ and $Sense_B$, where $Sense_A \in SynSet_A$ and $Sense_B \in SynSet_B$. 
In the prover, a user can restrict the number of word senses per word by specifying 
a cutoff $N$, i.e. the $N$ most frequent senses per word.

In addition to the WN relations, a user can introduce new lexical relations in the KB as Prolog facts, e.g., \verb+is_(crowd, group)+.

\comments{
the  there are ca. 90 relations manually added to the KB. 
A few of them concern quantifiers, e.g., 
$\synt{most} \subs \synt{a}$ and 
$\synt{few} \bot \synt{many}$, 
while the rest is the facts that could not be extracted from WordNet, e.g.,
$\synt{polish} \subs \synt{clean}$, 
$\synt{fight} \subs \synt{match}$, and
$\synt{crowd} \subs \synt{group}$. 
\addlater{Instance relation?}
}

\subsection{The Proof Engine}
\label{sub:engine}

The proof engine (PE) is the component that builds proof trees.
While applying rules it takes into account computational efficiency of each rule where the efficiency depends on the following categories:

\begin{itemize}[leftmargin=*]\itemsep0mm

\item {\it Branching}: a rule is either branching (e.g., $\forall_\T$) or non-branching (e.g., $\textsc{aux}$). 

\item {\it Semantic equivalence}: this depends whether the antecedents of a rule is semantically equivalent to its consequents.
For example, ($\wedge_F$) encodes the semantic equivalence while ($\textsc{no}_\T$) does not.

\item {\it Producing}: depending on whether a rule produces a fresh entity, it is a producer or a non-producer.
($\exists_\T$) is a producer while ($\forall_\T$) is not.

\item {\it Consuming}: a rule is a consumer iff it employs an old entity from the branch during application.
The consumer rules are ($\forall_\T$) and ($\textsc{no}_\T$) but ($\exists_\T$).

\end{itemize}

The most efficient combination of these features is non-branching, semantic equivalence, non-producing and non-consuming.
Depending on a priority order between these categories, called an efficiency criterion, one can define a partial efficiency order over the rules.
In particular, \eqref{ec:ebpc} is one of the best efficiency criteria on SICK \cite[Ch.\,6]{abzianidzethesis}. 
According to \eqref{ec:ebpc}, ($\wedge_\F$) is more efficient than ($\textsc{no}^n_\T$) since the equivalence is the most prominent category in \eqref{ec:ebpc}, and ($\wedge_\F$) is equivalence in contrast to ($\textsc{no}^n_\T$)
\begingroup
\abovedisplayskip=2mm\abovedisplayshortskip=1mm
\belowdisplayskip=2mm\belowdisplayshortskip=1mm
\begin{align}
[\semt{equi}, \semt{nonBr}, \semt{nonProd}, \semt{nonCons}]
\label{ec:ebpc}
\end{align} 
\endgroup  
A user can change the default criterion \eqref{ec:ebpc} by passing a criterion via the Prolog predicate ${\tt effCr/1}$.

The PE builds two structures: a tree (see Figure\,\ref{fig:pug}) and a list.
The latter represents a list of the tree branches.
The list structure is the main data structure that guides the computation process while the tree structure is optional (activated with the predicate ${\tt prooftree/0}$) and is used for displaying proofs in a compact way. 
A few of the predicates that control the proof procedure are:

\begin{itemize}[leftmargin=*]\itemsep0mm

\item ${\tt ral/1}$ sets a rule application limit to $n$, which means that after $n$ rules are applied the proof is terminated.
$n=400$ by default.

\item ${\tt thE/0}$ always permits existential import from definite NPs: it makes $(\exists_\T)$ applicable to the entry $\synt{the}_{\nou,\vp,\sen}\, \synt{dog}_\nou\, \synt{bark}_\vp:\F$. 

\item ${\tt allInt/0}$ allows to treat lexical modifiers of the form $c^{\tt VB.|JJ|NN}_{\nou,\nou}$ as intersective by default unless stated differently in the SG.
This permits to infer $\synt{baby}_\nou:c:\T$ and $\synt{kangaroo}_\nou:c:\T$ from $\synt{baby}_{\nou,\nou}\synt{kangaroo}:c:\T$, for better or worse.    

\item ${\tt the/0}$, ${\tt a2the/0}$, and ${\tt s2the/0}$ are used as flags and treat bare, indefinite, and plural NPs as definite NPs, respectively. 
\end{itemize}

\section{\langpro{}: Natural Language Prover}
\label{sec:langpro}

The tableau-based theorem prover for natural language is obtained by chaining a CCG parser, \llfgen{} and \nlogpro{}.
In order to detect a semantic relation between a set of premises $\{p_i\}_{i=1}^n$ and a hypothesis $h$, first the corresponding LLFs $\{P_i\}_{i=1}^n$ and $H$ are obtained via a CCG parser and \llfgen{} (i.e. for simplicity, a single LLF per sentence).
Then based on the lexical terms of the LLFs, relevant sets of relations $K$ and rules $R$ are collected from the KB and the IR, respectively.
To refute both entailment and contradiction relations \nlogpro{} builds two proof trees using $K$ and $R$. 
One starts with the counterexample \eqref{ent_tab} for entailment and another with the counterexample \eqref{cont_tab} for contradiction.
The semantic relation which could not be refuted (i.e. its tableau for the counterexample was closed) is said to be proved. 
The relation is considered to be neutral iff both tableaux have the same closure status: open or closed.%
%
%
\begingroup
\abovedisplayskip=0mm\abovedisplayshortskip=0mm
\belowdisplayskip=0mm\belowdisplayshortskip=0mm
\begin{align}
\{P_1:\T, ~~ \ldots, ~~ P_n:\T, ~~~ H:\F\}
\label{ent_tab}
\\
\{P_1:\T, ~ \ldots, ~ P_n:\T, ~~~ H:\T\}
\label{cont_tab}
\end{align} 
\endgroup

Entailment relations often do not depend on semantics of phrases shared by premises and hypotheses.
To bypass analyzing the common phrases, \langpro{} can use an optional CCG term aligner in \llfgen{} (Figure\,\ref{fig:langPro_lean}), which identifies the common CCG sub-terms and treats them as constants. 
The sub-terms that are downward monotone or indefinite NPs are excluded from alignments as they do not behave semantically as constants.   
After aligning CCG terms, aligned LLFs are obtained from them via the type-raising.
Tableau proofs with aligned LLFs are shorter.
Thus, first, a tableau with aligned LLFs is built, and if the tableau did not close, then non-aligned LLFs are used since alignment might prevent the tableau from closing.    
On SICK, the aligner boosts the accuracy by 1\%.
If stronger alignment is used (i.e. aligning indefinite NPs), the accuracy on SICK is increased by 2\%.
Both weak and strong alignment options can be chosen in \langpro{}.


The parser component of \langpro{} can be filled by C\&C or EasyCCG.
This results in two versions of \langpro{}, cc\langpro{} and easy\langpro{} respectively.
Both versions achieve similar results on FraCaS and SICK, and a simple aggregation of their judgments (co\langpro{}) improves the accuracy on the unseen portion of SICK by 1\%.

With respect to its rule-based nature, \langpro{} is fast.
Given ready CCG derivations, 
on average 100 SICK problems are classified in 3.5 seconds.%
\footnote{This is measured on $8 \times 2.4$ GHz CPU machine, when proving problems in parallel (via the $\mathtt{paralle/0}$ predicate) with the strong aligner option and the rule application limit $50$---the configuration that achieves high performance both in terms of speed and accuracy.
} 
Details about speed and impact of parameters on the performance are given in \newcite{abzianidzethesis}.  

In addition to an entailment judgment, \langpro{} can output the actual tableau proof trees (similar to Figure\,\ref{fig:pug}) in three formats: a drawing of a proof tree via the XPCE GUI, a \LaTeX{} source code, an XML output, or an HTML file.

\section{Related work}
\label{sec:related}

Theorem proving techniques \cite{BosMarkert2005EMNLP} or ideas from Natural Logic \cite{maccartneythesis} were already used in recognizing textual entailment (RTE).
But the combination of these two is a novel approach to RTE.
The underlying higher-order logic of \langpro{} guarantees sound reasoning over several premises, including some complex semantic phenomena.
This is in contrast to the RTE systems that cannot reason over several premises or cannot account for Booleans and quantifiers, including the ones \cite{maccartneythesis} inspired by Natural Logic, and in contrast to those ones that use FOL representations 
and cannot cover higher-order phenomena like generalized quantifiers or subsective adjectives.

\langpro{} achieves state-of-the-art semantic competence (with accuracy of 87\%) on the FraCaS sections commonly used for evaluation \cite{abzianidze:2016:*SEM,abzianidzethesis}.   
On SICK, the prover obtains 82.1\% of accuracy \cite{abzianidze:2015:EMNLP,abzianidzethesis} while state-of-the-art systems score in the range of 81-87\% and average performance of human on the dataset is around 84\%.  
Detailed comparison of \langpro{} to the related RTE systems is discussed in \cite{abzianidze:2015:EMNLP,abzianidze:2016:*SEM,abzianidzethesis}.


\section{Conclusion}
\label{sec:concl}


The presented natural language prover involves a unique combination of natural logic, higher-order logic and a tableau method.
Its natural logic side simplifies generation of the logical forms and makes the prover to be relatively easily scaled up.
Due to its higher-order virtue, the prover easily accounts for complex semantic phenomena untameable in FOL.
Because of its high reliability (less than 3\% of its entailment and contradiction judgments are incorrect), the judgments of the prover can be successfully borrowed by other RTE systems. 
Further scaling-up for longer sentences (e.g., newswire text) and automated knowledge acquisition present future challenges to the prover.

\section*{Acknowledgments}

This work has been supported by the NWO-VICI grant
``Lost in Translation -- Found in Meaning'' (288-89-003).

\bibliography{langpro}
\bibliographystyle{emnlp_natbib}

\end{document}